\documentclass{article}

    \usepackage[preprint]{neurips_2020}

\usepackage[utf8]{inputenc} 
\usepackage[T1]{fontenc}    
\usepackage{hyperref}       
\usepackage{url}            
\usepackage{booktabs}       
\usepackage{amsfonts}       
\usepackage{nicefrac}       
\usepackage{microtype}      
\usepackage{graphicx}       

\title{Conceptualization and Framework of Hybrid Intelligence Systems}

\author{%
  Nikhil ~Prakash\\
  \texttt{nikhil07prakash@gmail.com} \\
   \And
   Kory W. Mathewson \\
   \texttt{korymath@gmail.com} \\
}

\begin{document}

\maketitle

\begin{abstract}
As artificial intelligence (AI) systems are getting ubiquitous within our society, issues related to its fairness, accountability, and transparency are increasing rapidly. As a result, researchers are integrating humans with AI systems to build robust and reliable hybrid intelligence systems. However, a proper conceptualization of these systems does not underpin this rapid growth. This article provides a precise definition of hybrid intelligence systems as well as explains its relation with other similar concepts through our proposed framework and examples from contemporary literature. The framework breakdowns the relationship between a human and a machine in terms of the degree of coupling and the directive authority of each party. Finally, we argue that all AI systems are hybrid intelligence systems, so human factors need to be examined at every stage of such systems' lifecycle.
\end{abstract}

\section{Introduction}

Artificial intelligence (AI) systems are moving out of the research laboratories into the real world. Systems are deployed in critical sectors such as medicine, judiciary, and transportation, as a result, its limitations are becoming profound and leading to disastrous consequences.\footnote{Uber's self-driving operator charged over fatal crash (https://bbc.in/36knk3F)} To overcome these shortcomings, researchers have begun to explore various ways in which humans can be integrated with AI systems. This recent surge has led to the development of multiple hybrid intelligence systems (\citealt{evorus, calendar, scribe}). In a survey on the applications of crowdsourcing for machine learning research, \citealt{crowd} compiled a list of hybrid intelligence systems, from both the research community and industry, to explain the diversity in their use cases.

Although there has been a great deal of interest in the application of hybrid intelligence systems, little work has focused on its conceptual understanding, especially in relation to other similar concepts such as \textit{interactive machine learning}, \textit{AI-infused systems}, and \textit{mixed-initiative interactions}. This lack of conceptualization and diversity in terminology not only impedes the progress of the field but also makes it difficult for AI practitioners to make efficient design decisions.

In this work, we propose a precise definition of hybrid intelligence systems and describe its proper interpretation using examples. To explain its relation with other similar concepts and study new upcoming hybrid intelligence systems in the context of older ones, we propose an intuitive framework and report its usage by explaining a few well-known intelligent systems. Finally, we argue, based on our proposed definition, that all AI systems are hybrid intelligence systems and discuss the importance of recognizing human factors at every stage of such systems' lifecycle.

\section{Definition of Hybrid Intelligence Systems}
\label{section2}


Before providing our definition of hybrid intelligence systems, we first explain two other types of intelligent systems, \textit{Human Computation} and \textit{Self-Sufficient Artificial Intelligence} systems, and their relation with hybrid intelligence systems. The reason behind this is two-fold: it will establish proper boundaries and help to provide a broader yet concrete definition of hybrid intelligence systems. In this article, we use the definition of intelligence proposed by \citealt{intelligence}, ``\textit{as the ability of a system to act appropriately in an uncertain environment, where appropriate action is that which increases the probability of success, and success is the achievement of behavioral subgoals that support the system’s ultimate goal}.''

\subsection{Human Computation}

Although the term human computation has been used in multiple disciplines like philosophy and psychology for a very long time, its use in computer science is relatively recent (\citealt{HumanComp}). According to \citealt{HumanComp}, von Ahn's 2005 dissertation titled ``Human Computation'' was pivotal in instigating its usage in modern computer science literature, which defines it as ``\textit{a paradigm for utilizing human processing power to solve problems that computers cannot yet solve}.'' Consequently, the systems which solely utilize human computation, during its entire lifecycle, to complete its tasks are defined as human computation systems.

While in such systems, tasks are completed entirely through human intelligence, machines or software without intelligence could be utilized to assist humans. For instance, \textit{Heteroglossia} is a crowd-powered system that assists story writers by eliciting novel story ideas from the crowd workers (\citealt{story}). The system utilizes software applications such as Google Docs's comment functionality for streamline interaction, but it does not involve any AI in supporting the user. Hence, it is a human computation system. Similarly, 
Mechanical Turk\footnote{https://en.wikipedia.org/wiki/The\_Turk} can also be considered as a human computation system because the actual task of playing chess was performed by a human, and the machinery used to assist the human did not possess any intelligence. However, if there was any intelligence-possessing functionality in the machinery, then it would be regarded as a hybrid intelligence system.


\subsection{Self-Sufficient Artificial Intelligence}

Artificial Intelligence (AI) refers to the intelligence possessed by machines, thus sometimes also known as machine intelligence. Unlike humans, machines acquire intelligence through algorithmic techniques inspired from domains like statistics, mathematical optimization, cognitive science, and fueled by computer processing power and a large amount of data.

We define Self-Sufficient Artificial Intelligence systems as a type of intelligent system that does not utilize human computation at any point in its life cycle, i.e. human assistance is needed neither during its development nor when it is operational. The definition makes it evident that such intelligent systems could only be developed by either other self-sufficient artificial intelligence or hybrid intelligence systems. A key point to note here is that the self-sufficient artificial intelligence system created by a hybrid intelligence system is not considered a hybrid intelligence system because there is no direct involvement of human intelligence, and as a result, it will have inherently different characteristics. For instance, we don't have to consider human factors for the designing and development of such systems. Although we have not been able to find any such system, it is conceivable that large and complex hybrid intelligence systems would be able to create other intelligent systems, at least simpler than itself, without any direct human assistance in the near future, especially considering the recent developments in the field of Evolutionary Computation (\citealt{nature}, \citealt{neuro}, \citealt{evoAutoML}). While in automatic machine learning (AutoML) an intelligent system creates another one, it still requires data generated through human intelligence (\citealt{autoML1}). Therefore, AutoML systems are not self-sufficient artificial intelligence systems.

\subsection{Hybrid Intelligence}

Hybrid intelligence bridges the enormous gap between human computation and self-sufficient artificial intelligence and is defined as the paradigm which utilizes both human and machine intelligence to solve problems. The systems utilizing hybrid intelligence at any point in their life cycle are called hybrid intelligence systems. Except for human computation and self-sufficient artificial intelligence systems, the proposed concept of hybrid intelligence systems forms the superset of all other kinds of intelligent systems. The following set of intelligent systems is not intended to be exhaustive, but to provide an idea of how the proposed definition should be interpreted.

\paragraph{Interactive Machine Learning systems} 
Intelligent systems that are developed through rapid, focused, and incremental human interactions are considered as interactive machine learning systems (\citealt{IML}). There is a direct interaction between human and machine intelligence during the training phase in these systems, thus making them a special kind of hybrid intelligence system.

\paragraph{Machine / Deep Learning systems}
Although there is no direct interaction between human and machine intelligence in these systems, a considerable amount of human computation is still utilized at various stages such as data generation, feature extraction, result interpretation, and debugging. Thus, these systems also form a subset of the hybrid intelligence system.

\paragraph{AI-infused systems}
According to \citealt{guidelines}, systems that have features harnessing AI capabilities that are directly exposed to the end-user are referred to as AI-infused systems. In these intelligent systems, human computation is required not only during its development stages but also during its operations. Hence, these are a special kind of hybrid intelligence system as well.

\section{Framework}
The definition of hybrid intelligence systems provided in the previous section is broad and includes many different types of intelligent systems. As a result, it makes it difficult for AI researchers and practitioners to study and analyze various subsets of hybrid intelligence systems independently that share a number of common characteristics. Such targeted investigation could fast-track the progress of the field. Therefore, in this section, we explain our intuitive yet robust framework to overcome this challenge.

\begin{figure}[t]
    \centering
    \includegraphics[width=\linewidth, height=9cm]{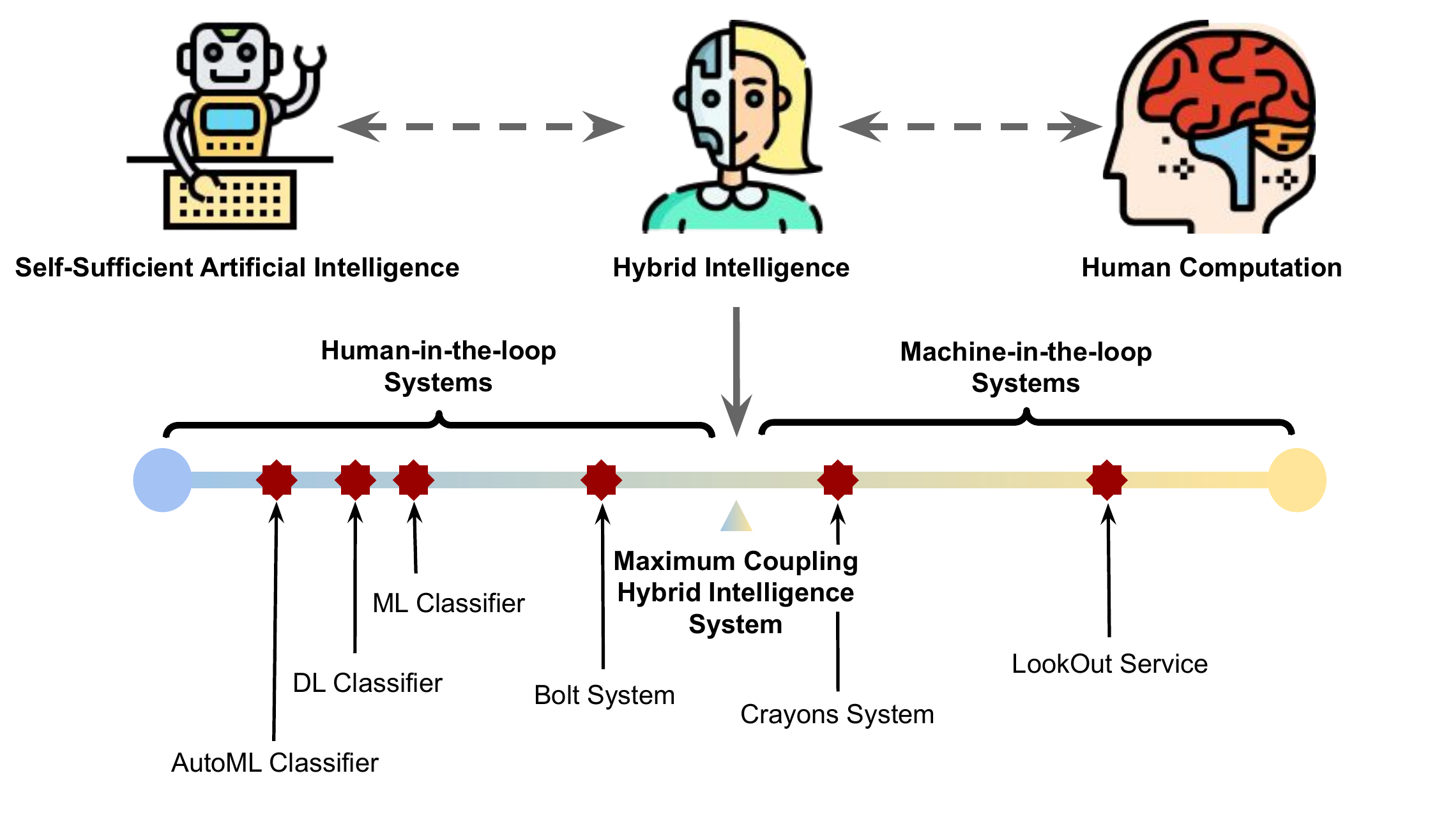}
    \caption{The continuum of hybrid intelligence systems with machine-in-the-loop and human-in-the-loop systems halves. Individual points on the continuum represent distinct hybrid intelligence systems with varying level of coupling between human and machine intelligence.}
    \label{fig1}
\end{figure}

As shown in the figure \ref{fig1}, the domain of hybrid intelligence systems is conceptualized as a continuum and divided into two halves, representing \textit{human-in-the-loop} and \textit{ machine-in-the-loop} systems.

\paragraph{Machine-in-the-loop systems}
These are the systems that utilize AI, to assist humans in solving their tasks. In this type of systems, the dominant component is human, and the minor component is AI. The endpoint on this half represents a hybrid intelligence system that has minimum AI and maximum human intelligence component.

\paragraph{Human-in-the-loop systems}
These are the systems in which human computation is utilized for assisting AI at any point in its life cycle. In this type of systems, the dominant component is AI, and the minor component is human. The endpoint on this half represents a hybrid intelligence system that has minimum human intelligence and maximum AI component.

The dimensions used to position hybrid intelligence systems on the continuum are:
\begin{enumerate}
    \item The coupling between human and machine intelligence. By coupling, we mean the level of integration between both the intelligence. It represents the distance from the center of the continuum.
    
    \item Who is playing the role of the dominant component? It represents the direction.
\end{enumerate}

The center of the continuum represents the highest level of coupling between human and machine intelligence. Systems at the center have an equal contribution of human and machine intelligence, and both are highly dependent on each other. As the coupling decreases the position of the system moves away from the center and the direction is determined by the dominant component. If the dominant component is human, then the system will shift towards machine-in-the-loop systems half and human-in-the-loop systems half otherwise.

The following list of hybrid intelligence systems is not intended to be exhaustive, but to provide an idea of how the proposed framework should be utilized.

\paragraph{LookOut Service} It is integrated with Microsoft Outlook and employs an intelligent agent to assist its users in reviewing their calendars and composing appointments when a new email is brought to focus (\citealt{mixed}). This service is considered a hybrid intelligence system because it uses both machine intelligence (intelligent agent) and human intelligence (end-user) to complete the task of composing appointments. The dominant component is human as it makes the final decision using the suggestion made by the agent. The interdependence between both the intelligence is relatively low, thus making it a loosely coupled hybrid intelligence system. Appropriating both the dimensions, it is positioned on the machine-in-the-loop systems half, as shown in figure \ref{fig1}.

\paragraph{Crayons System}  \citealt{fails} proposed an interactive machine learning system to train a pixel classifier, called Crayons system. People iteratively provided input to the system through brushstrokes by delineating the foreground and background of the image that was utilized to train the model. Based on the user studies conducted to evaluate the system, the authors demonstrated that human input was influenced by the machine output. Thus, making both the involved intelligence in the system highly coupled. Although in tightly coupled hybrid intelligence systems, the boundary between the dominant and minor component becomes uncertain, we argue that in this system, human intelligence is the dominant component as it has control over the training process. Hence, this system is positioned near the center of the continuum on the machine-in-the-loop systems half, as shown in figure \ref{fig1}.

\paragraph{Bolt System} \citealt{bolt} introduced the idea of \textit{instantaneous crowdsourcing systems}: intelligent systems that present human computation at machine-level speed, typically in the order of milliseconds. They proposed a hybrid intelligence workflow called the \textit{look-ahead approach} to develop such systems, which is based on the idea of prefetching crowd workers' response to the probable future states of the system. To demonstrate its efficacy, they built a hybrid intelligence system called the \textit{Bolt} system, modeled on the Markov Decision Process to predict the upcoming states. For each predicted state, the system queried and fetched crowd worker's responses. When the system moves to any of these states, the prefetched responses are used to provide instantaneous actions. In this system, both intelligence contribute significantly to determine its output, thus coupling between intelligence is high. Further, we argue that AI is playing the role of the dominant component as it controls which states need to be annotated and in what order by the crowd workers. Hence, the system is positioned near the center of the continuum on the human-in-the-loop half.

\paragraph{Cat Image Classifier} In a typical machine learning (ML) workflow for developing a cat image classifier, human intelligence is required at multiple stages like data collection, pre-processing, model design, and debugging, which makes it a hybrid intelligence system. In these systems, humans provide their assistance to develop the machine learning model, which performs the actual classification task. Hence, AI is considered as the dominant component, and the coupling between both intelligence is relatively low. Consequently, this system is positioned on the human-in-the-loop systems half on the continuum. Feature engineering is excluded in the deep learning (DL) workflow, which reduces the involvement of human intelligence, decreasing the integration between both intelligence. With the advent of AutoML, the requirement for human intelligence in the workflow has been significantly reduced, further decreasing the coupling between both the intelligence. This gradual reduction of coupling is represented as the system being shifted towards the endpoint of the human-in-the-loop systems in figure \ref{fig1}.

\paragraph{Evolving Hybrid Intelligence Systems} Intelligent systems that evolve throughout their lifecycle, i.e. the contribution of human and machine intelligence change over time, can easily be explained through our proposed framework. One such system is \textit{Calendar.help}, proposed by \citealt{calendar}. It is a service that allows a user to schedule meetings conveniently and provide an experience similar to that of working with a human assistant. The user emails (carbon copy) the system's virtual assistant to use the service. After receiving the email, the virtual assistant follows up with the other person to decide upon the details, creates the meeting, and sends the invitation automatically. The proposed system's architecture was developed through multiple design iterations. Starting as a human computation system, it utilized domain experts to perform its tasks. In the next iteration, some of the tasks were transferred to crowdworkers. In the final version, machine intelligence was introduced to automate a few tasks. Authors have also noted that the dependence on human intelligence would be further reduced as more data is collected. This gradual progression during design iterations can be easily explained through our proposed framework as moving the system from human computation to tightly coupled hybrid intelligence and eventually to the human-in-the-loop systems endpoint. However, such a system could never be categorized as a self-sufficient artificial intelligence system, even if it could operate independently after some time because a considerable amount of human computation has gone into its development.

The exact unit of coupling between both intelligence, as well as examples that exist at the endpoints or the center of the continuum, have not been intentionally provided, as we believe that such statements can be made only when the field has matured and saturation is achieved concerning various types of hybrid intelligence systems. It is difficult to assert what kind of hybrid intelligence systems would have a maximum and minimum coupling of human and machine intelligence at this point. Therefore, these crucial types of hybrid intelligence systems can be thought of as theoretical limits.

\section{Discussion}

As mentioned in section \ref{section2}, our proposed definition of hybrid intelligence systems is quite broad and forms a superset of all other kinds of intelligent systems. It also makes it evident that indeed all AI systems, except the self-sufficient artificial intelligence systems, are hybrid intelligence systems. And we believe that it is an important step towards acknowledging the much-needed fact that humans are an integral part of any AI workflow. It will compel designers, developers, and deployers to think about humans at every step in the process and obey the standard guidelines and principles in the domain (\citealt{guidelines, humancentered}). Such practices would encourage the development of human-centered intelligent systems that put humans before machines and help mitigate problems related to fairness, accountability, transparency, and ethics. More generally, this work aims to instigate more discussion on how AI practitioners could be encouraged and incentivized to give higher importance to human factors when designing, developing, and deploying intelligent systems.

Our proposed framework can help researchers to organize, group, and classify different types of existing and upcoming intelligent systems. It would lead to a better understanding of existing terminologies and bring consistency in the field. Developers and practitioners will also benefit from the organization as it will help them to make a better system design decision based on their requirements. It can also assist policymakers by targeting various types of hybrid intelligence systems to build better governance mechanisms.

\section{Related Work}

\citealt{HIS} provided an initial conceptualization of the term hybrid intelligence systems by defining it as ``\textit{systems that have the ability to accomplish complex goals by combining human and artificial intelligence to collectively achieve superior results than each of them could have done in separation and continuously improve by learning from each other}''. In this definition, the condition of continuous learning by both humans and machines restricts its applicability to most of the intelligent systems, which utilize both human and machine intelligence. For example, Teachable Machine,\footnote{https://teachablemachine.withgoogle.com/} an interactive machine learning system in which the end-user trains a classifier through direct interaction (iteratively provides input and evaluates output), will not be considered as a hybrid intelligence system as the human does not learn anything new by teaching the model.  

\citealt{prosthetic} proposed that humans as well as prosthetics and other technologies for human enhancement should be treated as autonomous agents of an information processing system. They provided a schema to analyze the relationship between the ability of each agent to have, seek, and achieve its goals and the overall human-machine performance. Based on the schema and existing works in the literature, they concluded that overall human-machine performance can be significantly improved by increasing the capabilities of prosthetic agents. 

\citealt{direction} studied the reasoning capabilities of AI systems to determine when and how should it query humans to maximize its benefit from human computation, with a focus to enhance its applicability. \citealt{hyints} has postulated that hybrid intelligence systems could bridge the socio-technical gap between the capabilities of systems and varying demands of people. \citealt{crowdplatforms} also addressed the challenges and opportunities for developing better crowdsourcing platforms with the rise of hybrid intelligence systems.

\section{Conclusion and Future Work}

In this work, we provided a precise definition of hybrid intelligence systems and argued that it forms a superset of all other similar concepts such as interactive machine learning, AI-infused systems, and mixed-initiative interactions. We also provided a simple intuitive framework to compare and contrast existing and upcoming hybrid intelligence systems based on the degree of coupling between human and machine intelligence and the directive authority of each party, along with examples from the literature. It would be interesting to investigate various properties, strengths, and limitations of systems positioned at different locations on the continuum. It would also be worth investing time to investigate the exact unit of coupling and precise definition of the maximum and minimum coupling between human and machine intelligence with examples.

\section*{Broader Impact}

This article presents theoretical work on the precise definition and intuitive framework of hybrid intelligence systems to underpin the rapid growth in the development of such systems. It investigates the interaction of human and machine intelligence to build accountable, reliable, and trustworthy intelligent systems. It also advocates the idea of examining human factors at every stage of the intelligent systems' lifecycle. This research will not only benefit researchers but also AI practitioners in the industry. This work does not include any system building and is not related to any particular dataset. We believe that this work has no adverse impact on society.

\begin{ack}
We thank anonymous reviewers for their helpful suggestions and constructive feedback.
\end{ack}

\bibliographystyle{abbrvnat}
\bibliography{sample}

\end{document}